\documentclass[lettersize,journal]{IEEEtran}
\usepackage{amsmath,amsfonts}
\usepackage{array}
\usepackage[caption=false,font=normalsize,labelfont=sf,textfont=sf]{subfig}
\usepackage{textcomp}
\usepackage{stfloats}
\usepackage{url}
\usepackage{verbatim}
\usepackage{graphicx}
\usepackage{cite}

\usepackage{graphicx}
\usepackage{booktabs}

\usepackage{algorithm2e}
\usepackage{algpseudocode}
\usepackage{array}

\usepackage{pifont}
\newcommand{\cmark}{\ding{51}}
\newcommand{\xmark}{\ding{55}}

\usepackage{hyperref}

\begin{document}

\title{Competing for pixels: a self-play algorithm for weakly-supervised segmentation}

\author{Shaheer U. Saeed,
        Shiqi Huang,
        Jo\~ao Ramalhinho,
        Iani J.M.B. Gayo,
        Nina Monta\~na-Brown,
        Ester Bonmati,\\
        Stephen P. Pereira, 
        Brian Davidson, 
        Dean C. Barratt,
        Matthew J. Clarkson, 
        Yipeng Hu
\IEEEcompsocitemizethanks{
\IEEEcompsocthanksitem S.U. Saeed, S. Huang, J. Ramalhinho, I.J.M.B. Gayo, N. Monta\~na-Brown, E. Bonmati, D.C. Barratt, M.J. Clarkson and Y. Hu are with
the Centre for Medical Image Computing, Wellcome/EPSRC Centre
for Interventional and Surgical Sciences, Department of Medical
Physics and Biomedical Engineering, University College London,
London WC1E 6BT, U.K.
\IEEEcompsocthanksitem S. Huang is also with the School of Optics and Photonics, Beijing Institute of Technology, Beijing, China.
\IEEEcompsocthanksitem E. Bonmati is also with the School of Computer Science and Engineering, University of Westminster, London W1W 6UW, U.K.
 \IEEEcompsocthanksitem S.P. Pereira is with the Institute for Liver and Digestive
Health, University College London, London NW3 2QG, U.K.
\IEEEcompsocthanksitem B. Davidson is with the Division of Surgery and Interventional Sciences, University College London, London WC1E 6BT, U.K.
\\
\IEEEcompsocthanksitem E-mail: shaheer.saeed.17@ucl.ac.uk
}%
}


\markboth{Saeed et al. 2024. Competing for pixels: a self-play algorithm for weakly supervised segmentation.}%
{Saeed et al. 2024. Competing for pixels: a self-play algorithm for weakly supervised segmentation.}


\maketitle

\begin{abstract}
Weakly-supervised segmentation (WSS) methods, reliant on image-level labels indicating object presence, lack explicit correspondence between labels and regions of interest (ROIs), posing a significant challenge. 
Despite this, WSS methods have attracted attention due to their much lower annotation costs compared to fully-supervised segmentation. Leveraging reinforcement learning (RL) self-play, we propose a novel WSS method that gamifies image segmentation of a ROI.
We formulate segmentation as a competition between two agents that compete to select ROI-containing patches until exhaustion of all such patches. The score at each time-step, used to compute the reward for agent training, represents likelihood of object presence within the selection, determined by an object presence detector pre-trained using only image-level binary classification labels of object presence. 
Additionally, we propose a game termination condition that can be called by either side upon exhaustion of all ROI-containing patches, followed by the selection of a final patch from each.
Upon termination, the agent is incentivised if ROI-containing patches are exhausted or disincentivised if an ROI-containing patch is found by the competitor. This competitive setup ensures minimisation of over- or under-segmentation, a common problem with WSS methods. Extensive experimentation across four datasets demonstrates significant performance improvements over recent state-of-the-art methods.
\\
Code: \url{https://github.com/s-sd/spurl/tree/main/wss}
\end{abstract}

\begin{IEEEkeywords}
Self-Play, Weak Supervision, Segmentation.
\end{IEEEkeywords}

\section{Introduction}\label{sec:intro}

\begin{figure}
    \centering
    \includegraphics[width=0.48\textwidth]{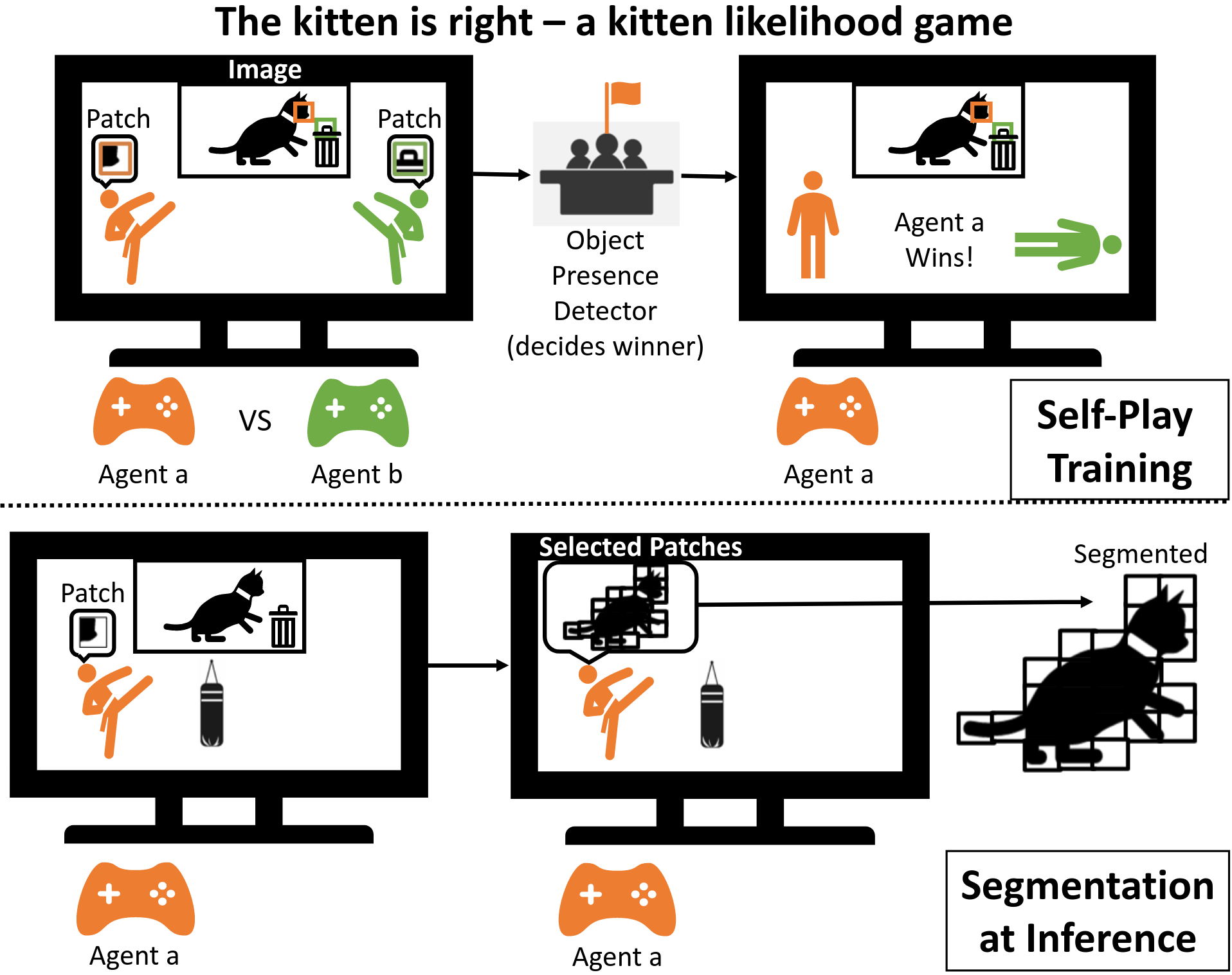}
    \caption{Competitive self-play for segmentation using only weak signals generated by an object presence detector (itself trained using only image-level classification labels). At inference the trained agent plays against a dummy opponent to segment the image.}
    \label{fig:motivation}
\end{figure}

Segmentation is a fundamental task in computer vision with demonstrated applications in e.g., medical image processing \cite{sharma2010automated, ramesh2021review}, autonomous driving \cite{feng2020deep}, and robotics \cite{minaee2021image}. Fully supervised segmentation systems, trained using human-annotated data, have widely been proposed, however, they are limited by the expensive pixel-level annotation acquisition for datasets of increasing sizes \cite{minaee2021image}. Tasks or datasets, where annotation cost is high or where these labels are not readily available, are especially hindered by this requirement \cite{minaee2021image} e.g., in the medical domain where expert time may be restricted. Weakly supervised segmentation (WSS) aims to directly address these challenges by utilising weak labels for training, such as image-level classification labels (i.e., binary labels of object presence) \cite{zhou2018brief}. 

We propose a WSS method in which we gamify segmentation such that two opposing agents compete to segment a region-of-interest (ROI), where each move is scored by a weak signal derived from a pre-trained object presence detector. This detector is a classifier of object presence, trained only using image-level classification labels. This weak signal is similar to classifiers used in recent WSS methods \cite{zhou2018brief} and ensures discriminative features, able to distinguish object existence from non-existence, are learnt only using image-level labels. The framework is based on reinforcement learning (RL) self-play where the competitive setup encourages minimisation of over- or under-segmentation, a common problem with WSS \cite{chen2022self, zhou2022regional}. 

In our framework, an image is first divided into patches using a grid of a pre-defined size, where the patches are akin to playing-cards in a card game. An agent from each competing side (two sides in this case) selects, without replacement, a patch that it deems most likely to contain a part of an ROI. The game is turn-based and thus prevents selection of the same patch by opposing sides. Each patch is scored by the pre-trained object presence detector, using the unbounded classification probability, or logits, and the agent with a higher score wins the turn and is given a positive reward signal. This is repeated until one of the agents delivers a termination signal, which indicates when an agent considers all ROI-containing patches to be exhausted. The final patches from each side are then scored by the object presence detector and if the patch from the side that delivered the terminal signal is scored higher, then that side wins the game - leading to a high positive reward signal. The main idea is summarised in Fig. \ref{fig:motivation}. This may be similar to an innate self-play-style approach within the human brain, where object presence likelihood for parts of the field-of-view (FOV) may be decided before localisation \cite{yun2013exploring}.

Learning to maximise the score/ reward using RL, in this competitive setup, ensures that under-segmentation is minimised as it is not beneficial to terminate the game early. This is because early termination may mean that the opponent could potentially select a more object-like patch, or the opponent's selected patch may be scored higher simply based on chance e.g., if one part of the object is more object-like than the other.
It also ensures that over-segmentation is minimised, as timely termination is rewarded, once ROI-containing patches are exhausted. If the game is not terminated appropriately and both sides keep selecting patches with no ROIs, the opposing side may win the game based on chance as one part of the background may be more object-like compared to another. Over thousands of games, these early or later termination cases, lead to reduced expectation for the reward, compared to an effective termination strategy which terminates only when ROI-containing patches are exhausted. Thus RL, which learns by maximising the expectation of reward learns an effective termination strategy. In practice, limiting numbers of turns (by discouraging termination beyond the specified range) and selecting patch sizes sufficiently smaller than approximate ROI size were sufficient constraints to learn effective segmentation. Once trained, an agent may select patches without opposition to get a segmentation with resolution equal to patch-size.

Using self-play not only improves performance, by ensuring that over- or under-segmentation are minimised through an effective termination strategy, but also proves more time-efficient compared to scoring all patches using the object presence detector as non-selected patches do not need to pass through the detector. It also avoids the need for empirical thresholds for classification probability for patch-based segmentation. If pixel-level segmentation is required, the grid can be shifted over by half the patch-size and the game repeated, where the final segmentation may be obtained using a majority vote. This still proves more time-efficient compared to a sliding windows approach where the object-presence detector must score all patches. 

\subsection{Summary of contributions}

1) We propose a gamified competitive self-play framework for image segmentation which minimises over- or under-segmentation, a common problem for other WSS methods; 2) We propose to train this WSS framework using RL where the reward is based solely on an object presence detector trained only using image-level classification labels of object presence; 3) We evaluate our proposed approach using two benchmarks datasets for WSS (PASCAL VOC 2012 and MS COCO 2014) as well as two real-world medical datasets, representing real-world applications where pixel-level human annotations are expensive to acquire and curate (liver tumour and prostate gland segmentation).

\section{Related work}

\subsection{Reinforcement learning self-play}

Self-play describes a special class of multi-agent problems in RL where two or more agents compete to accomplish a task, often set-up as a zero-sum game \cite{digiovanni2021survey, shoham2003multi}. In self-play, an agent competes with itself, or a copy of itself, to improve without the need for any direct supervision e.g., human labels. Rather, a reward signal, often based on the win/ loss conditions of a game, guides learning where both copies of the agent compete to maximise the reward, or in other words to win the game. This promotes learning competitive behaviours such as developing counters to effective opposing high-reward strategies or developing defensive strategies \cite{digiovanni2021survey, silver2017mastering, silver2018general}. Perhaps, the main benefit of self-play comes from acquiring multiple sets of experiences that can be used to train a single agent \cite{silver2017mastering, silver2018general, heinrich2015fictitious, heinrich2016deep}. This not only promotes faster learning compared to learning against a hand-crafted strategy, but also allows higher diversity in experiences for potentially improved generalisability \cite{silver2017mastering, silver2018general, bai2020near, bai2020provable, heinrich2016deep, heinrich2015fictitious}. Self-play also aids learning by allowing a near-perfect adversary, since the agent plays against an opponent that is equally skilled, a copy of itself, and improves as the opponent improves \cite{laterre2018ranked}. 

Thus far, the focus for self-play research has mostly been to use learning in the context of existing zero-sum games e.g., Chess \cite{silver2018general}, Go \cite{silver2017mastering}, Poker \cite{heinrich2016deep} and Hide and Seek \cite{baker2019emergent}. These games and self-play research in general can be classified by information-completeness based on the comparative information accessible to each of the players \cite{digiovanni2021survey, silver2017mastering, silver2018general}. 
Chess, where both players have the same information, and are fully aware of the opposition's payoffs and actions, are known as information-complete. Any games where an aspect of the game dynamics is hidden from any of the players are known as information-incomplete e.g., Hold'em Poker where the player is unaware of the observed state (card selection) for opposing players. Self-play has been extensively validated for both types of games. 

Roots of self-play are grounded in game theory, however, RL self-play was popularised by works that proposed general algorithms that enabled learning in information-complete environments such as Go \cite{silver2017mastering, silver2018general} and Othello \cite{van2013reinforcement}. They showed, for the first time, that RL self-play starting from random play, could outperform handcrafted strategies and sophisticated search techniques without any domain knowledge, except game rules. These techniques were built upon and applied to various complex information-complete computer games such as Tennis and Soccer \cite{liu2021self}.

Concurrently, research into using RL self-play for information-incomplete games also gained popularity through its use in games e.g., Poker \cite{heinrich2015fictitious, heinrich2016deep}. Further developments led to the application and extension of RL self-play to more challenging environments such as Capture the Flag \cite{team2021open}, DouDizhu \cite{zha2021douzero} and Robotic Soccer \cite{haarnoja2023learning}. 

The goal with self-play applied to any game or competitive task is to reach a Nash equilibrium, whereby an agent will learn a perfect policy from which it will not deviate \cite{shamma2005dynamic, portelas2020automatic, heinrich2015fictitious, heinrich2016deep, silver2017mastering, silver2018general, bai2020near}. However, in practice, reaching Nash equilibrium is often not feasible due to the large action and observation spaces in modern games, unless human knowledge is used to abstract domains to manageable sizes \cite{heinrich2016deep}. Nonetheless self-play using RL (without injecting any expert knowledge) has managed to learn near-optimal solutions to classical information-complete games \cite{silver2017mastering, silver2018general}. Several opponent sampling curricula have been proposed for efficient learning in an attempt to get closer to Nash equilibria in more complex environments.

In `vanilla self-play', often described as the `best response' curriculum, the current most up-to-date version of the agent is used as an opponent. This opponent has been trained on the most experience against itself, and thus may offer the best response to an average response of the opposing player \cite{heinrich2016deep}. This insight prompted the use of a mixture of average response and best response as an opponent instead of only the best response, as using only marginally weaker opponents may promote learning \cite{laterre2018ranked, vinyals2019grandmaster, brown1951iterative, leslie2006generalised, heinrich2016deep, portelas2020automatic}. The key insight from practical experiments being that a new mixed curriculum may lead to more generalisable agents due to the larger breadth of collected experience and also to avoidance of cyclic learning and local optimas common in vanilla self-play, where improvements are not made 
\cite{vinyals2019grandmaster, heinrich2015fictitious, brown1951iterative, leslie2006generalised, shamma2005dynamic}. This method of learning with an opponent with mixed responses (mixture of best and average responses) is known as `fictitious self-play'. In recent practice, in RL self-play, this may be implemented by randomly sampling past versions of agents (including the current agent itself) to act as opponents \cite{portelas2020automatic, vinyals2019grandmaster, liu2021self, wang2023self}. This holds theoretically as an expectation of randomly sampling all past responses is equivalent to the average response. Any curriculum that uses non-uniform sampling distributions over all past versions of the agent may be classed as `prioritised fictitious self-play' \cite{vinyals2019grandmaster}.  This often leads to faster learning, dependant on the sampling distribution, as only limited amounts of experience are collected against very weak opponents i.e., experience collection time is not wasted on much weaker opponents that pose no challenge to the current agent \cite{vinyals2019grandmaster}. Developments in competitor sampling have enabled various real-world applications.

In recent years, self-play research has focused on finding competitive gamified formulations for a variety problems (that are not pre-existing games) to find effective novel solutions in various domains. Kajiura et al. \cite{kajiura2020self} demonstrated the use of RL self-play in image re-targeting by proposing a competition between agents to select optimal image operators, to adjust images to arbitrary sizes without information loss, within reasonable times. Self-play found effective strategies, with processing times three orders of magnitude less than previous multi-operator methods. Hammar et al. \cite{hammar2020finding} showed that by gamifying infrastructure security attacks e.g., denial of service and cross-site scripting, and defence, e.g., firewalls, effective intrusion prevention strategies, that outperform tested baselines, could be learnt with RL fictitious self-play without expert knowledge. Recently, Wang et al. \cite{wang2023self} proposed RL prioritised fictitious self-play for protein engineering to mutate sequences towards desired properties, similar to playing pieces in a game, which led to the discovery of proteins with 7.8-fold bio-luminescence improvement compared to naturally occurring sequences.

\subsection{Weakly supervised segmentation}\label{sec:wss}

Instead of full pixel-level annotations, WSS utilises weak labels to guide segmentation, e.g., image-level \cite{araslanov2020single, jiang2019integral, wei2018revisiting}, patch-level \cite{dai2015boxsup, lee2021bbam, song2019box, oh2021background, li2020deep} or pointers and sparse free-hand annotations \cite{liang2022tree, lin2016scribblesup, vernaza2017learning, roth2021going}. In this work, we focus on weak signals derived from image-level labels as they are often most time-efficient to obtain and are already available in many real-world applications e.g., medical imaging diagnostic tasks \cite{zhou2018brief}. Furthermore, many works assume image-level labels as a pre-requisite for freehand-, or patch-based WSS, because labelling e.g., patches or points, automatically leads to a binary annotation of object presence \cite{zhou2018brief}. Missing image-level classification to ROI correspondence in WSS, in ground truth data, remains a challenging task to overcome with previously proposed solutions being based mainly on class activation maps (CAM) \cite{araslanov2020single, jiang2019integral, wei2018revisiting, li2021group}, or multi-instance learning (MIL) \cite{li2023weakly, jia2017constrained, xu2019camel, xu2014weakly}.

Methods for WSS based on CAM, extract ROIs from internal activations of a classification neural network, trained to classify object presence \cite{selvaraju2017grad}. These methods, however, may be prone to under-segment since classifiers are driven to activate only small portions of the image with strong discriminative capabilities \cite{zhou2022regional, chen2022self}. Solutions employ a number of techniques. Region growing extrapolates the initial map, with or without external information \cite{huang2018weakly, kolesnikov2016seed, wang2018weakly}. The so-called erasure-based methods remove highly discriminative features from images such that classifiers activate other less discriminative features \cite{hou2018self, kumar2017hide, lee2019ficklenet, wei2017object, wei2018revisiting, kweon2021unlocking, sun2021ecs, choe2020attention, zhang2020reliability}; erasure area, number of erasures, are all hyperparameters to be tuned, often per image. Background modelling creates maps of the background with the help of external saliency maps \cite{kim2021discriminative, lee2021railroad, xu2021leveraging, yao2021non} or using a class-wise discriminator re-trained for every class \cite{fan2020learning}. Information aggregation  accumulates information from multiple images to build object representations \cite{zhou2022regional, fan2020cian, liu2020weakly, sun2020mining}. Self-supervised learning uses auxiliary tasks to build supervisory signals \cite{chen2022self,chang2020weakly, shimoda2019self, wang2018weakly}. However, a general solution to WSS using only image-level classification labels remains an active research area.

MIL has been proposed for WSS due its ability to capture correspondence between bags of multiple instances, with available annotations, and single instances, to-be-labelled \cite{foulds2010review}. For WSS, works aim to pose images as bags with image-level classification labels available and pixels as instances, to be labelled \cite{jia2017constrained, li2023weakly}. Some solutions define bags as patches with labels available \cite{han2022multi}. Others incorporate clustering \cite{xu2014weakly}, or redefine instances as lattice patches to obtain coarse segmentation \cite{xu2019camel}, which may increase inference time due to the need to infer for all patches. Due to a focus on discriminative features as in CAM, possibly sufficient for classification but under-specific for full-object segmentation, these formulations also tend to under-segment.

\section{Methods}

In this work, we propose a gamified formulation for WSS where two agents compete to select ROI-containing patches until exhaustion of all such patches. The game score at each time-step is determined by an object presence detector which indicates the likelihood of a patch containing the ROI. The game score is used as a reward signal to train the agents using RL, and the object presence detector is pre-trained using only image-level object presence classification labels. The game may be terminated by either of the competing agents, once they consider all ROI-containing patches to be exhausted. If, post-termination, the opposing side is able to select a patch with higher likelihood of object presence (determined by the object presence detector), a negative reward signal is delivered to the side that called the termination signal, otherwise a positive reward is delivered. This competitive set-up ensures minimisation of over- or under-segmentation, as described in Sec. \ref{sec:intro} and outlined in the following subsections.

This formulation involves two types of functions: 1) object presence detector - a classifier trained using binary image-level classification labels of object presence, to classify object presence within an image or patch; and 2) the agent - responsible for patch selection such that a segmentation map may be generated, trained using RL self-play with rewards derived from the trained object presence detector. We first formalise our proposed prioritised fictitious self-play and then describe the set-up for WSS.

\begin{figure*}[!ht]
    \centering
    \includegraphics[width=0.98\textwidth]{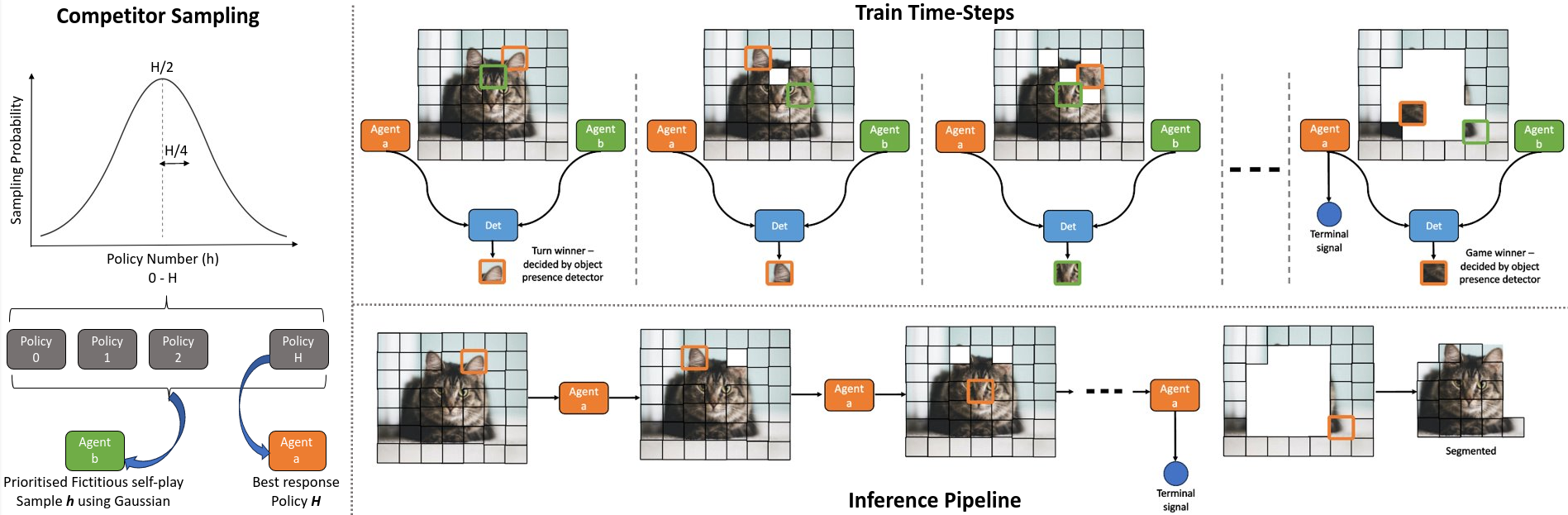}
    \caption{Overview of the gamified WSS segmentation using RL self-play.}
    \label{fig:self_play}
\end{figure*}

\subsection{Competing functions and the environment}

In the form of self-play of interest in this work, two opposing agents compete to achieve a goal in a zero-sum game. The problem formulation is described in this section. 

    \noindent\textbf{Single agent reinforcement learning:}
    RL involves learning optimal policies for decision-making under Markov decision process (MDP) environments. The environment may be defined as $(\mathcal{S}, \mathcal{A}, p, r, \pi, \gamma)$.

    \textbf{States, actions and the policy:} An agent or policy, interacting with an environment, from time-step $t$, $\pi(\cdot; \theta): \mathcal{S} \times \mathcal{A} \rightarrow \{0, 1\}$ represents the probability of performing action $a_t \in \mathcal{A}$ to influence the environment, given observed state $s_t \in \mathcal{S}$, where $\mathcal{A}$ and $\mathcal{S}$ are the action and state spaces, respectively, and the weights are $\theta$. An action is sampled using $a_t \sim \pi(\cdot| s_t; \theta)$. The state transition distribution conditioned on state-action pairs is $p: \mathcal{S} \times \mathcal{S} \times \mathcal{A} \rightarrow [0, 1]$, where probability of the next state $s_{t+1}$ given the current state-action pair $s_t, a_t$ is $p(s_{t+1} | s_t, a_t)$. 

    \textbf{Rewards:} The reward function, which indicates the value of performing a given action when within a given current state, is defined as $r: \mathcal{S} \times \mathcal{A} \rightarrow \mathbb{R}$, which, at step $t$, produces a reward $R_t = r(s_t, a_t)$.

    \textbf{Trajectories:} Following the state transition distribution for sampling next states, the policy for sampling actions and the reward function for sampling rewards, we can collect a trajectory of state-action-reward triplets over multiple time-steps $\tau = (s_1, a_1, R_1, \ldots, s_T, a_T, R_T)$. 

    \textbf{Policy optimisation:} The policy $\pi(\cdot; \theta)$, modelled as a neural network, with parameters $\theta$, predicts parameters of a categorical distribution from which an action is to be sampled $a_t \sim \pi(\cdot| s_t; \theta)$. For a policy followed to collect a trajectory, return is given by $R(\tau) = \sum_{k=0}^T \gamma^k R_{t+k}$, which indicates return over trajectory $\tau$. $\gamma$ is a discount factor for future rewards. The central optimisation problem to find optimal policy parameters $\theta^*$ then becomes:

    \begin{align}
    \theta^* = \arg \max_\theta \mathop{\mathbb{E}}_{\tau\sim\pi(\cdot; \theta)} [R(\tau)]
    \end{align}

    Expectation over trajectories is computed using multiple trajectories from agent-environment interactions, where $\tau\sim\pi(\cdot; \theta)$ denotes trajectory sampling using policy $\pi$.

    \noindent\textbf{Multi-agent self-play environments:}
    In self-play, if observed states for the two competing agents are of the same form, from the point-of-view of the competitors, then sampling an action for an observed state for the competitor may also be achieved using  $a_t \sim \pi(\cdot | s_t; \theta)$. 

    At this point, we introduce subscripts $t_a$ and $t_b$ where $a$ and $b$ indicate the agent $a$ or the competitor agent $b$, respectively. In this case, the states, actions and rewards for agent $a$ may be denoted as $s_{t_a}$, $a_{t_a}$, $R_{t_a}$ and those for agent $b$ as $s_{t_b}$, $a_{t_b}$, $R_{t_b}$. Trajectories for agents $a$ and $b$ are denoted as $\tau_a$ and $\tau_b$. The optimisation problem is then modified:

    \begin{align}\label{eq:selfplay_policy_update}
    \theta^* = \arg \max_\theta \{\mathop{\mathbb{E}}_{\tau_a \sim\pi(\cdot; \theta)} [R(\tau_a)] + \mathop{\mathbb{E}}_{\tau_b \sim\pi(\cdot; \theta)} [R(\tau_b)]\}
    \end{align}

    After each termination, two trajectories $\tau_a$ and $\tau_b$ are collected, one from each side. Using experience from both sides leads to a diverse range of experiences (from both the winning and losing side), potentially aiding generalisability. 


    \noindent\textbf{Competitor sampling:}
    While, the section above describes vanilla self-play, fictitious self-play may use a historic version of the policy denoted $\pi_h(\cdot; \theta_h)$, where $h$ denotes the training iteration from which the historic policy is sampled, when the current iteration is denoted as $H$, i.e., $h\in\{0, 1, 2, \ldots, H\}$, where $H$ is incremented by one on each training iteration. So agent $a$ is always denoted as $\pi_H(\cdot; \theta_H)$, using the most up-to-date policy or best response. 
    
    For fictitious self-play, the competitor agent $b$ is denoted as $\pi_h(\cdot; \theta_h)$, where $h\sim\rho(h)$ and $\rho(h) = 1 / H$ specifying a uniform distribution between $0$ to $H$, over all historic policies. That is to say that a competitor is chosen randomly from all past versions of the policy. 
    
    For prioritised fictitious self-play, we propose to use a truncated normal distribution for competitor sampling from historic policies. The competitor agent $b$ is denoted as $\pi_h(\cdot; \theta_h)$, where $h\sim\rho(h)$, where $\rho(h)$ takes the form:

    \begin{align}\label{eq:prioritised_fictitious_sampling}
        \rho(h) = \frac{4}{H} ~ \cdot ~ &\frac{ \psi\left( \frac{4h - 2H}{H} \right) }{ \Phi\left( -1 \right) - \Phi\left( -2 \right)}
    \end{align}

    \noindent Where $\psi(v) = \frac{1}{\sqrt{2\pi}} e^{-\frac{1}{2} v^2}$;
    and $\Phi(v) = \frac{1}{2} (1 + \text{erf}(v/\sqrt{2}))$;
    and $\text{erf}(v)$ is the Gauss error function.

    This is our proposed prioritised fictitious self-play. The distribution that we use is more commonly referred to as a truncated Gaussian distribution, where we use $H/2$ as the mean, $H/4$ as the standard deviation and $0$ and $H$ as the bounds. 
    This strategy is likely to select opponents that are weaker than the agent but still allows much weaker or the best response to be sampled for a wide range of experiences to be collected. At the same time, a large amount of collection time is not dedicated to much weaker opponents as in fictitious self-play and cyclic learning / local optima, encountered in vanilla self-play, are avoided as collection time for nearly equally skilled opponents is also reduced. 

    \noindent\textbf{Agent optimisation:}
    Optimisation for both fictitious or prioritised fictitious self-play are the same as vanilla self-play, with the only difference being that for vanilla self-play, for trajectory $\tau_a$ and $\tau_b$ collection, the actions for both agent $a$ and agent $b$ are sampled using $a_{t_a} \sim \pi_H(\cdot | s_{t_a}; \theta_H)$ and $a_{t_b} \sim \pi_H(\cdot | s_{t_b}; \theta_H)$, respectively (i.e., using the most recent or best-response policy), with states for each still being sampled using the state transition distribution $p$ and rewards for each also still computed using the reward function $r$. For fictitious or prioritised fictitious self-play, actions for agent $a$ and agent $b$ are sampled using $a_{t_a} \sim \pi_H(\cdot | s_{t_a}; \theta_H)$ and $a_{t_b} \sim \pi_h(\cdot | s_{t_b}; \theta_h)$, respectively (i.e., using the most recent or best-response policy for agent $a$ and sampling a competitor agent $b$ from historic policies), where $h\sim\rho(h)$, and $\rho(h)$ is defined based on whether self-play is fictitious when a uniform distribution over historic policies is used, or prioritised fictitious when a non-uniform distribution over historic policies is used. 

\subsection{Gamified segmentation using self-play}

\begin{algorithm}[!ht]
\caption{RL self-play for WSS.}
\label{algo:seg_self_play}
\SetAlgoLined
\KwData{Image-label pairs) $\{x_i, y_i\}_{i=1}^N$}
\KwResult{Trained RL policy $\pi(\cdot; \theta^*)$}
\BlankLine
Train object presence detector using image-label pairs (eq. \ref{eq:train_obj_pres_det}) to get $f(\cdot; w^*)$\;
Set current policy number $H=0$
\BlankLine
\While{not converged}{
Randomly sample an image $x_i$\;
Start at $t = 0$\;
Divide image $x_i$ into patches $\{ x_{i, p} \}_{p=1}^P$\;
Set agent $a$ to have policy $\pi_H(\cdot; \theta_H)$\;
Sample policy $\pi_h(\cdot; \theta_h)$ for agent $b$ (eq. \ref{eq:prioritised_fictitious_sampling})\;
Set the initial state as $s_0 = \{ x_{i, p} \}_{p=1}^P$\;
Sample action $a_{0_a} \sim \pi_H(\cdot| s_0)$ for agent $a$\;
Select patch $x_{i, p_a, 0}$ and erase to update $s_0$\;
Sample action $a_{0_b} \sim \pi_h(\cdot| s_0)$ for agent $b$\;
Select patch $x_{i, p_b, 0}$ and erase to update $s_0$\;
Compute reward $R_{0_a}$ for agent $a$ (eq. \ref{eq:fin_reward})\;
Compute reward $R_{0_b}$ for agent $b$ (eq. \ref{eq:fin_reward} with inverted inequalities)\;
\BlankLine
\For{$t\leftarrow 1$ \KwTo $t_{\text{max}}$}{
Carry $s_t = s_{t-1}$ from previous iteration \;
Sample action $a_{t_a} \sim \pi_H(\cdot| s_t)$ for agent $a$\;
Select patch $x_{i, p_a, t}$ and erase to update $s_t$\;
Sample action $a_{t_b} \sim \pi_h(\cdot| s_t)$ for agent $b$\;
Select patch $x_{i, p_b, t}$ and erase to update $s_t$\;
Compute reward $R_{t_a}$ for agent $a$ (eq. \ref{eq:fin_reward})\;
Compute reward $R_{t_b}$ for agent $b$ (eq. \ref{eq:fin_reward} with inverted inequalities)\;
Break iterating $t$ if $a_{t, \text{term}} = 1$; set $t=t_{\text{end}}$\;
}
\BlankLine
Once $R_{t_a=1:t_{\text{end}}}$ and $R_{t_b=1:t_{\text{end}}}$ collected, update policy using gradient ascent (eq. \ref{eq:selfplay_policy_update})\;
Update current policy number $H = H+1$}
\end{algorithm}

    \noindent\textbf{An overview of gamified segmentation:}
    Gamified segmentation consists of two agents that select patches from within an image, such that all ROI-containing patches are selected, where the object presence detector acts as a mechanism to score the patch selection by opposing agents. The score is based on the raw classification probability of the patch, from the object presence detector, which serves as a proxy for likelihood of ROI presence within the patch. The game is terminated by means of a termination signal delivered by one of the agents, when it deems all ROI-containing patches to be exhausted. Rewards are given for each move, where the winning patch at each step is determined by the object presence detector, as well as at termination to determine the winning side based on if the opponent can present a patch with a higher score or not, as determined by the object presence detector.

    \noindent\textbf{Object presence detector:}
    The object presence detector $f(\cdot; w): \mathcal{X} \rightarrow \left[0, 1\right]$ predicts object or ROI presence (zero for absence and one for presence) within an image or image patch $x \in \mathcal{X}$, where $\mathcal{X}$ is the domain of images and $w$ are the weights. The only training data used to train $f$ are image-label pairs $\{x_i, y_i\}_{i=1}^N$. Binary cross-entropy loss is used for training: $L_f(y_i,  f(x_i; w)) = ~(y_i) \log(f(x_i; w)) - (1 - y_i)\log(1 - f(x_i; w))$. And the optimisation problem is to find optimal weights $w^*$, by means of gradient descent:
    
    \begin{align}\label{eq:train_obj_pres_det}
    w^* = \arg \min_w \mathop{\mathbb{E}} \left[L_f(y, f(x; w))\right]    
    \end{align}
    
    After training, the classifier predicts raw classification probability for the binary object presence classification task $f(x_i; w^*) = \left[0, 1\right]$ for image $x_i$, which is a proxy for likelihood of object presence. Note that while training, only images and image-level classification labels of object presence are used, however, inference may be on image patches that can be interpolated to the image size used for training, regardless of patch size. This enables quantification of the likelihood of object presence within an image patch, which we will use as a reward generator to train our RL self-play segmentation framework, as outlined in the following sections. This kind of task-based reward for RL is inspired by previous works \cite{yoon2020data, saeed2022image, saeed2022image_2, cubuk2018autoaugment, zhang2019adversarial, saeed2021adaptable, saeed2021learning}.
    
    \noindent\textbf{The environment:}
    The gamified segmentation environment encapsulates the image to-be-segmented $x_i\in\mathcal{X}$ and the object presence detector $f(\cdot; w^*)$ with pre-trained weights $w^*$. All states, actions and rewards are defined from the perspective of agent $a$, however, definitions may be adapted to agent $b$ e.g., by inverting inequalities defined in rewards. Subscripts $a$ are omitted from some terms for readability, however, are added where ambiguous or where both agents $a$ and $b$ are involved. A single time-step $t$ encompasses moves from both sides, i.e., one patch selection action each (see Algo. \ref{algo:seg_self_play}).

    \textbf{States:} The observed state for the environment is an image $x_i$ divided into $P$ patches using a grid of pre-defined size, $s_{t} = \{ x_{i, p} \}_{p=1}^P$, where $x_{i,p}$ is a patch of image $x_i$. 

    \textbf{Actions:} We use a $P$-dimensional discrete action space for patch selection $a_{t, \text{patch}} \in \{1, 2, \ldots, P\}$, which selects one of $P$ patches, and a 2-dimensional discrete action space for the termination action $a_{t, \text{term}} \in\{0, 1\}$ (one for termination and zero otherwise). The final action space is then given by $a_t = \left[a_{t, \text{patch}}, a_{t, \text{term}}\right]^\top \in \mathcal{A}$. The patch number selected by agent $a$ is defined as $p_a$ and for agent $b$ as $p_b$, otherwise a subscript of $a$ or $b$ at the action, state or reward indicates the respective agent that is responsible.

    \textbf{Patch selection} At any time-step $t$, a patch selected by agent $a$ is denoted as $x_{i,p_a,t}$, which once selected, is replaced by a zero-intensity patch (pixel values equal to zero) to update observed state $s_t$, i.e., $x_{i,p_a,t} = \mathbf{O}$. This is similar to commonly-used erasure-based WSS works \cite{choe2020attention, yi2023boundary, hou2018self, kumar2017hide, lee2019ficklenet, huang2021flip}, such that selecting that patch again is not advantageous in terms of score, since the object presence detector is likely to produce a low score for any zero-intensity patch, in addition to an explicit reward dis-encouraging such selections (see below). The observed state is modified twice at each time-step, once by each of the two agents. After this erasure of selected patches, the updated state is used as the next state $s_{t+1}$.

    \textbf{Rewards:} Reward at each time-step has four parts: 
    
    1) the patch reward:

    $
    R_{t, \text{patch}} = \begin{cases} 
      +1 & \textbf{if} ~ f(x_{i, p_a, t}; w^*) \geq f(x_{i, p_b, t}; w^*)\\
      -1 & \text{otherwise}\\
   \end{cases}
    $

    Which indicates the winning patch at each time-step, as scored by the object presence detector, where the higher raw classification probability for a patch wins the turn. Note that the game is sequential turn-based where one agent plays before another to prevent selection of the same patches by the two agents. This makes the setup a zero-sum game.

    2) the terminal reward:

$     R_{t, \text{term}} =    \begin{cases} 
    ~~~~ 0  ~ \textbf{if} ~ a_{t, \text{term}} = 0\\
    ~~~~~~ \textbf{otherwise:}\\
    ~~~~~~~ +1 ~ \textbf{if} ~ f(x_{i, p_a, t}; w^*) \geq f(x_{i, p_b, t}; w^*) \\
     ~~~~~~~ -1 ~ \textbf{if} ~ f(x_{i, p_a, t}; w^*) < f(x_{i, p_b, t}; w^*) \\
   \end{cases}
$

Which decides the final game winner at termination such that if a termination signal is delivered by agent $a$ and agent $a$ loses the turn, a negative reward is given otherwise if agent $a$ wins the terminal time-step, a positive reward is delivered. The terminal reward plays an important role in preventing over- and under- segmentation otherwise encountered with image-level supervision (see Sec. \ref{sec:wss} and \ref{sec:intro}).

    3) the repetition reward:

$    R_{t, \text{rep}} = \begin{cases}       -1 & \textbf{if} ~ x_{i,p_a,t} = \mathbf{O}\\
      ~~~0 & \text{otherwise}\\
   \end{cases}
$

Which discourages selection of an already selected patch as these are replaced by zero-intensity patches $\mathbf{O}$ in state $s_t$.

    4) the iteration bounding reward:

$    R_{t, \text{iter}} = \begin{cases} 
      -1 & \textbf{if} ~ t_{\text{min}} < t < t_{\text{max}} ~ \text{and} ~ a_{t, \text{term}} = 1\\
      ~~~0 & \text{otherwise}\\
   \end{cases}
$

    Which prevents early termination before $t_{\text{min}}$ and also late termination after $t_{\text{max}}$.
    
    In practice $R_{t, \text{rep}}$ and $R_{t, \text{iter}}$ act as shaping to make learning faster, however, are not necessary (see Table \ref{tab:res_abl_all}).

    The final reward for the time-step $t$ is thus formed using all four components in a weighted manner:
    
    \begin{align}\label{eq:fin_reward}
        R_t = R_{t, \text{patch}} + 100R_{t, \text{term}} + R_{t, \text{rep}} + R_{t, \text{iter}}
    \end{align}

    Where the weights were tuned on a small subset of the PASCAL VOC 2012 train set (see Sec. \ref{sec:datasets}).

    \noindent\textbf{Agent-environment interactions:}
    The above-described states, actions and rewards are used to train a policy (eq. \ref{eq:selfplay_policy_update}). An overview is outlined in Algo. \ref{algo:seg_self_play} and Fig. \ref{fig:self_play}. 
    At inference, we run the policy without opposition, or with a fixed opponent policy which always selects an artificially-placed zero-intensity patch. Inference to obtain a a final segmentation map is done via the agent selecting patches containing the object without opposition. The grid of patches can be shifted by one pixel, or more, in order to run inference again to conduct a majority vote over pixel selections to get a pixel-level segmentation map.

\section{Experiments and results}\label{sec:exp_res}

\begin{figure*}
    \centering
    \includegraphics[width=0.9\textwidth]{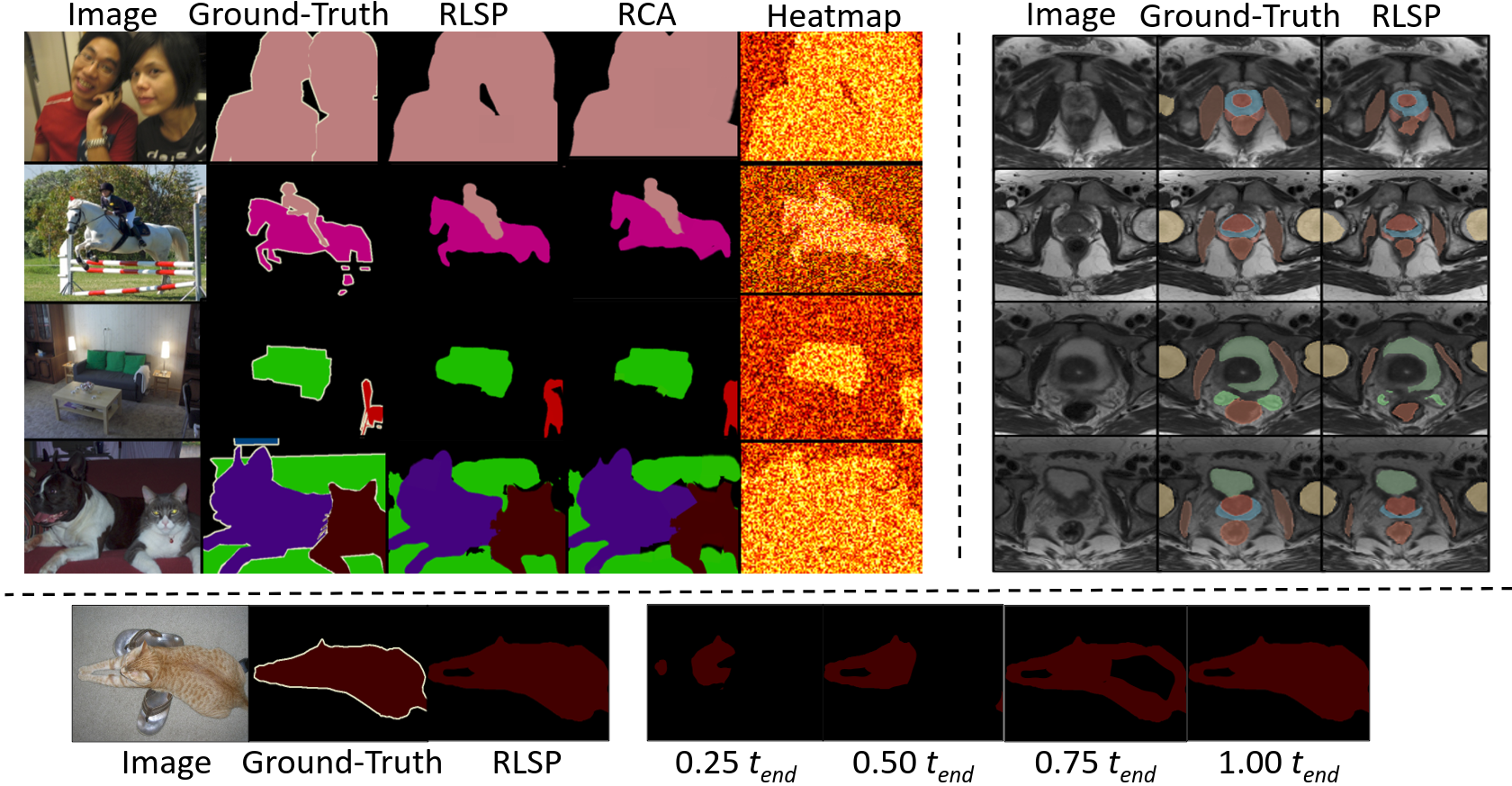}
    \caption{Left: VOC; Right: CMPS. 'Heatmap': object presence detector score heatmap. Bottom: VOC; time-points at inference. RLSP uses pixel-level majority vote by running inference thrice.}
    \label{fig:seg_samples}
\end{figure*}

\subsection{Datasets for evaluation}\label{sec:datasets}

\noindent\textbf{PASCAL VOC 2012 (VOC) \cite{everingham2015pascal}} has 20 classes with 1,464, 1,449 and 1,456 images in the train, validation and test sets. As common practice \cite{zhou2022regional, chen2022self, huang2018weakly}, the train set was augmented with 10,582 images \cite{hariharan2011semantic}, with 1024 used for monitoring performance, and the validation and test sets were held out and used to report results (test results obtained from evaluation server).

\noindent \textbf{MS COCO 2014 (COCO) \cite{lin2014microsoft}} has 80 classes with 80k and 40k images in the train and validation sets (20k images used from train set for monitoring performance). Validation set was held out and used to report results, as common practice \cite{chen2022self, zhou2022regional}.

\noindent \textbf{Male Pelvic Structures (CMPS) \cite{li2023prototypical}} has 8 classes with 471 and 118 samples in the train and test sets (64 samples from train set used for monitoring performance). Test set was held out and used to report performance. We use 2D slices with each 3D sample containing approximately 20-40 slices.

\noindent \textbf{Liver Tumour (LiTS) \cite{bilic2023liver}} has 1 class (tumour) with 107 and 24 samples in the train and test sets (24 images from train set used for monitoring performance). The test set was held out and used to report performance. We use 2D slices with each 3D sample containing approximately 200-400 slices.

\subsection{Network architectures and hyper-parameters}

We use ResNet38 \cite{he2016deep, wu2019wider} as a backbone for our object presence detector and agent. We train a separate network for identifying object presence or absence, for each class (threshold 0.5 used at inference, to decide classes present in the image). Hyper-parameter values are summarised in Tab. \ref{tab:hyp_obj_pre}.

\begin{table}[!ht]
\centering
\caption{Hyper-parameter values.}
\begin{tabular}{c|c}
\hline
Hyper-parameter & Value \\
\hline
\multicolumn{2}{c}{}\\
\hline
\multicolumn{2}{c}{Object presence detector}\\
\hline
Input-size & 256$\times$256\\
Architecture & ResNet38\\
Batch-size & 64\\
Optimiser & Adam\\
\hline
\multicolumn{2}{c}{}\\
\hline
\multicolumn{2}{c}{Self-play}\\
\hline
RL algorithm & Policy Gradient\\
Input-size & 256$\times$256\\
Architecture & ResNet38\\
Batch-size & 128\\
Optimiser & Adam\\
Gamma $\gamma$ & 0.96\\
$t_{\text{min}}$ & 4\\
$t_{\text{max}}$ & 1024\\
Patch-size & 6$\times$6\\
\hline
\end{tabular}
\label{tab:hyp_obj_pre}
\end{table}

\subsection{Comparisons with the state-of-the-art}

We compare our proposed RL self-play (RLSP) against multiple proposed CAM or MIL-based WSS works that utilise only image-level labels for training (citations provided where appropriate). Where we re-implemented a method this is denoted with a suffix `RE'.

Previous works do not report measures of spread so statistical tests were not conducted. For re-implemented methods, we report t-test results (significance $\alpha=0.05$). 

We report the commonly used overlap measures mean intersection over union (mIoU), for VOC and COCO, or mean binary Dice score (mDSC), for CMPS and LiTS, dependant on what compared methods have reported.

\newcolumntype{P}[1]{>{\centering\arraybackslash}p{#1}}

\begin{table*}[!ht]
\centering
\caption{Segmentation results. VOC, COCO (mIoU); CMPS, LiTS (mDSC).}
\begin{tabular}{c c P{1cm} P{1cm} P{1cm} P{1cm} P{1cm}}
\hline
Method & Backbone & VOC (val) & VOC (test) & COCO & CMPS (RE) & LiTS (RE) \\
\hline
SEC \cite{kolesnikov2016seed} & VGG16       & -    & -    & 22.4 & - & - \\
IRN \cite{ahn2019weakly} & ResNet50         & 63.5 & 64.8 & 32.6 & - & - \\
IAL \cite{wang2020weakly} & VGG16           & -    & -    & 27.7 & - & - \\
ICD \cite{fan2020learning} & ResNet101      & 64.1 & 64.3 & -    & - & - \\ 
SCE \cite{chang2020weakly} & ResNet101      & 66.1 & 65.9 & -    & - & - \\ 
SEAM \cite{wang2020self} & ResNet38         & 64.5 & 65.7 & 31.9 & - & - \\
BES \cite{chen2020weakly} & ResNet101       & 65.7 & 66.6 & -    & - & - \\
MCIS \cite{sun2020mining} & ResNet101       & 66.2 & 66.9 & -    & - & - \\
CONTA \cite{zhang2020causal}& ResNet101     & 66.1 & 66.7 & 33.4 & - & - \\
LIID \cite{liu2020leveraging} & ResNet101   & 66.5 & 67.5 & -    & - & - \\
A2GNN \cite{zhang2021affinity} & ResNet101  & 66.8 & 67.4 & -    & - & - \\
AdvCAM \cite{lee2021anti} & ResNet101       & 68.1 & 68.0 & -    & - & - \\
CDA \cite{su2021context} & ResNet38         & 66.1 & 66.8 & -    & - & - \\
ECS \cite{sun2021ecs} & ResNet38            & 66.6 & 67.6 & -    & - & - \\
CSE \cite{kweon2021unlocking} & ResNet38    & 68.4 & 68.2 & 36.4 & - & - \\
CPN \cite{zhang2021complementary} &ResNet38 & 67.8 & 68.5 & -    & - & - \\
USAGE \cite{peng2023usage} & ResNet38       & 71.9 & 72.8 & 44.3 & - & - \\
MIL (RE) \cite{li2023weakly} & VGG16        & 67.9 & - & 37.9 & 52.1 & 67.4 \\
RCA \cite{zhou2022regional} & ResNet38      & 72.2 & 72.8 & 36.8 & 53.9 & 68.8 \\
SIPE \cite{chen2022self} & ResNet38         & 68.2 & 69.5 & 43.6 & 53.7 & 66.1 \\
MCT \cite{chen2022self} & ResNet38         & 71.9 & 71.6 & 42.0 & 54.1 & 67.1 \\
ACR \cite{chen2022self} & ResNet38         & 72.4 & 71.9 & 45.3 & 53.7 & 67.9 \\
MARS \cite{jo2023mars} & ResNet101         & 77.7 & 77.2 & 49.4 & 54.2 & 68.6 \\
\hline
RLSP (ours) & ResNet38                     & 78.9 & 78.7 & 49.9 & 58.8 & 70.7 \\
\hline
\end{tabular}
\label{tab:res}
\end{table*}

Table \ref{tab:res} shows that our method outperformed recent state-of-the-art (SOTA) WSS methods that only use image-level classification labels for training, on all tested datasets. We observe improvements of 1.2 mIoU points on VOC (validation set), 1.7 mIoU points on VOC (test set), 0.5 mIoU points on COCO, 4.6 mDSC points on the CMPS, and 1.9 mDSC points on the LiTS dataset. For the CMPS and LiTS datasets, differences are statistically significant (p-values$<$0.001). Modifying the backbone of our method for VOC did not substantially impact performance i.e., changing from ResNet38 to ResNet101 or VGG16 marginally reduced performance, by 0.5 or 1.2 mIoU points, respectively.

\begin{figure}
    \centering
    \includegraphics[width=0.38\textwidth]{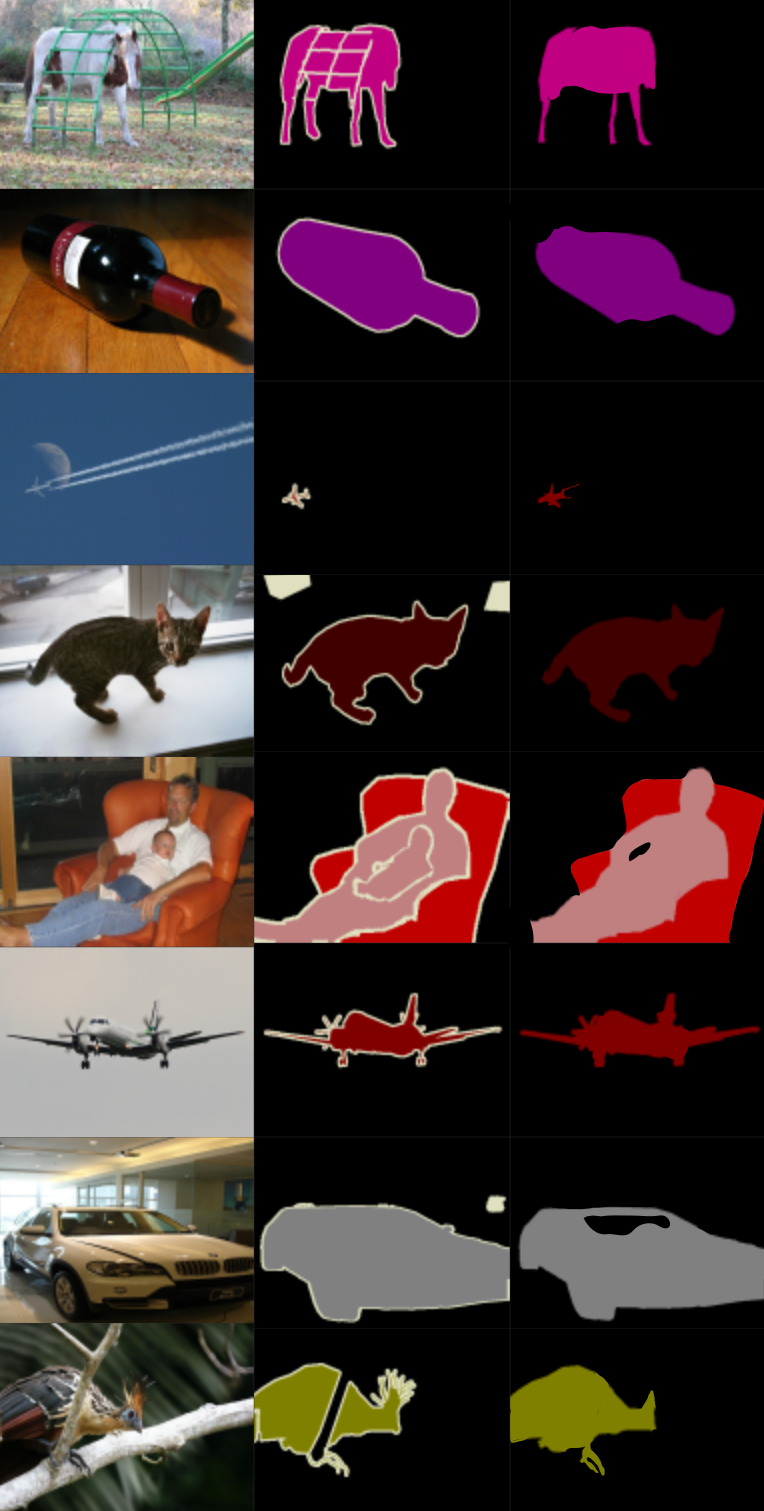}
    \caption{VOC test set segmentation results (left to right: image; ground truth; proposed).}
    \label{fig:seg_samples_additional}
\end{figure}

We observe that our proposed method performs best under scenarios where the ROI is smaller than (less than half) the whole FOV of the image (qualitative samples in Fig. \ref{fig:seg_samples} and \ref{fig:seg_samples_additional}).

Training and inference times varied with the number of classes and approximate ROI size in the dataset e.g., for VOC, we used approximately 450 GPU hours for training on an Nvidia Tesla V100 GPU (48 hours pre-training object presence detector), and 2.5 seconds for inference per image. Performance improvements across a wide variety of datasets demonstrate the applicability of our framework.

\subsection{Ablation studies}

Table \ref{tab:res_abl_patch_size} shows impact of increasing patch-size. This leads to improved performance, until a plateau, which is intuitive as our classifier tends to perform well when input FOV matches the training FOV i.e., as the input patches get larger and closer to whole images. Future development could explore an adaptive object presence detector that is trained together with the agents for optimal performance for a given patch size. However, since VOC had a wide variety of FOVs in the training data, impact of patch size was limited.

\begin{table}[!ht]
\centering
\caption{Results for varying patch size. VOC (val) (mIoU).}
\begin{tabular}{c c}
\hline
Patch size & VOC\\
\hline
2 $\times$ 2 & 76.4\\
4 $\times$ 4 & 76.6 \\
6 $\times$ 6 & 78.9 \\
8 $\times$ 8 & 78.3\\
10 $\times$ 10 & 78.4\\
12 $\times$ 12 & 78.8\\
16 $\times$ 16 & 78.7\\
\hline
\end{tabular}
\label{tab:res_abl_patch_size}
\end{table}

Table \ref{tab:res_abl_all}, shows impact of removal of components from our proposed framework. Non-self-play variants are trained with a reward based on a threshold 0.8 in place of agent $b$, i.e., setting $f(x_{i, p_b, t}; w^*) = 0.8$. Omission of shaping refers to repetition and/ or iteration bounding rewards i.e., $R_{t, \text{rep}}=0$ and/ or $R_{t, \text{iter}}=0$. We observe significant negative impact without self-play, however, impact is limited when shaping rewards are omitted. Note that reward shaping reduced training times by approximately 8 hours.

\begin{table}[!ht]
\centering
\caption{Results for ablation study. VOC (val) (mIoU).}
\begin{tabular}{c c c c}
\hline
SP & SR & VOC\\
    & $R_{t,\text{rep}}  ~|~ R_{t,\text{iter}}$ & \\ 
\hline
\xmark & \xmark ~~~~~~~ \xmark & 68.8\\
\xmark & \cmark ~~~~~~~ \cmark & 70.3\\
\cmark & \xmark ~~~~~~~ \xmark & 75.7\\
\cmark & \cmark ~~~~~~~ \xmark & 76.4\\
\cmark & \xmark ~~~~~~~ \cmark & 77.7\\
\cmark & \cmark ~~~~~~~ \cmark & 78.9\\
\hline
\end{tabular}
\label{tab:res_abl_all}
\end{table}

\subsection{Comparisons to baselines}

Table \ref{tab:baselines}, shows comparisons to various baseline models such as a fully-supervised network (trained with pixel-level ground truth), sliding window approach (using the object presence detector with a commonly used threshold of 0.5 for segmentation), and a single-agent formulation denoted as non-SP, with varying thresholds used for selecting patches. We demonstrate superior performance of our proposed framework compared to all tested variants trained with weak image-level labels. Furthermore, we report results for different seeds and show that our training scheme is robust to seed selection with a variance (standard deviation) of $0.47$ over 8 random seeds. Given that patch-size has limited impact and that compared to sliding window approach without any RL, we obtain a higher performance for RLSP, combined with results from the ablation study, we conclude that self-play is the largest positive contributor towards performance in our proposed framework.

\begin{table}[!ht]
\centering
\caption{Comparisons to baselines. VOC (val) (mIoU).}
\begin{tabular}{c c}
\hline
Variant & VOC\\
\hline
Fully-supervised                        & 81.2\\
Non-SP ($f(x_{i, p_b, t}; w^*) = 0.8$)  & 70.3\\
Non-SP ($f(x_{i, p_b, t}; w^*) = 0.5$)  & 69.8\\
Non-SP ($f(x_{i, p_b, t}; w^*) = 0.2$)  & 68.5\\
Sliding-window                          & 64.9\\
RLSP (seed 1)                           & 78.9\\
RLSP (seed 2)                           & 77.9\\
RLSP (seed 3)                           & 78.5\\
\hline
\end{tabular}
\label{tab:baselines}
\end{table}

Performance improvements in overlap measures (mIoU and mDSC) compared to other methods (see Table \ref{tab:res}), or variants (see Table \ref{tab:baselines}) with sliding windows or omitted self-play, indicate the ability of our proposed method to provide effective segmentation. These overlap measures are reduced by over- or under-segmentation and thus we conclude that our method, with the highest overlap measure amongst tested WSS methods, learns an effective termination strategy which minimises over- or under-segmentation compared to other methods and tested variants.

\subsection{Quantifying over- and under-segmentation}

Additionally, in Table \ref{tab:ratios} we quantify over- and under-segmentation by reporting the proportion of true-positive (TP), true-negative (TN), false-positive (FP) and false-negative (FN) pixel values, averaged over the set (normalised to a sum of 100). The higher TP and TN and lower FP and FN for our method further indicate that over- and under-segmentation were minimised compared to other tested methods. 

\begin{table}[!ht]
\centering
\caption{Comparing over-/ under-segmentation. VOC (val).}
\begin{tabular}{c c c c c c}
\hline
Method & mIoU & TP & TN & FP & FN \\
\hline
Sliding-window & 64.9 & 63 & 15 & 10 & 12 \\ 
Non-SP & 70.3 & 65 & 15 & 10 & 12 \\
MARS \cite{jo2023mars} & 77.7 & 67 & 17 & 8 & 8  \\
RLSP & 78.9 & 69 & 18 & 6 & 7 \\
\hline
\end{tabular}
\label{tab:ratios}
\end{table}


\section{Discussion and conclusion}




The results presented in Sec. \ref{sec:exp_res} demonstrate the effectiveness of our proposed method, trained only using weak image-level classification labels, by showing improved performance compared to competing methods, ablated variants and other baselines. The tasks on which our framework is evaluated demonstrate a wide range of applications including two multi-class segmentation computer-vision datasets and two medical imaging datasets, showing applicability to various types of images and ROIs.

Direct comparisons to other methods, in terms correct classifications of pixels containing ROIs, using true-positive, true-negative, false-positive and false-negative ratios, also demonstrate the ability of the method to reduce over- or under-segmentation as it shows higher true-negative and -positive rates and lower false-negative and -positive rates compared to other tested methods.

The task-based rewards, in the self-play set-up, proposed in this work, open up potential avenues for future work where weak supervision signals may be derived for tasks such as image-alignment, object localisation in 3D space, or anomaly detection. 
Moreover, extending our task-based rewards holds promise for facilitating meta-learning within a self-play context. In such set-ups two task networks may engage in competitive interactions supervised by an outer-level network, all trained simultaneously, enhancing the modelling of intricate interdependencies across these functions.
In conclusion, our novel self-play framework opens up further avenues for research into supervisory signals and meta-learning configurations.



In this work, we propose a gamified WSS such that competing agents segment an ROI from an image trying to minimise over- or under-segmentation. The weak supervision signals used during training are derived only from an object presence detector, which is able to classify object presence within a selected image patch, and itself is trained only using image-level binary classification labels of object presence. Extensive experiments on four datasets show that our proposed framework outperforms recently proposed SOTA methods on two well-known WSS benchmarks as well as on two real-world medical datasets.

\section*{Acknowledgements}

This work is supported by the EPSRC grant [EP/T029404/1]; Wellcome/EPSRC Centre for Interventional and Surgical Sciences [203145Z/16/Z]; the EPSRC CDT in i4health [EP/S021930/1]; and the International Alliance for Cancer Early Detection, an alliance between Cancer Research UK [C28070/A30912; C73666/A31378], Canary Center at Stanford University, the University of Cambridge, OHSU Knight Cancer Institute, University College London and the University of Manchester.

%
%
\bibliographystyle{splncs04}
\bibliography{bib}

\end{document}